%% file: main.tex
\def\year{2021}\relax
\documentclass[letterpaper]{article} 
\usepackage{aaai21}  
\usepackage{times}  
\usepackage{helvet} 
\usepackage{courier}  
\usepackage[hyphens]{url}  
\usepackage{graphicx} 
\usepackage{natbib}  
\usepackage{caption} 
\usepackage{url}            
\usepackage{amsfonts}       
\usepackage{xcolor}

\usepackage{multirow}
\usepackage{soul}
\usepackage{url}
\usepackage[utf8]{inputenc}
\usepackage{graphicx}
\usepackage{amsmath}
\usepackage{amsthm}
\usepackage{booktabs}
\usepackage{subcaption}
\usepackage[ruled]{algorithm2e}
\usepackage[switch]{lineno} 

\theoremstyle{defn}
\newtheorem{defn}{Definition}

\title{Multifaceted Context Representation using Dual Attention for Ontology Alignment}
\author{
        Vivek Iyer\textsuperscript{\rm{1}},  
    Arvind Agarwal\textsuperscript{\rm{2}},
    Harshit Kumar\textsuperscript{\rm{2}}
    \\
}
\affiliations{

    \textsuperscript{\rm 1}IIIT Hyderabad, India,
    vivek.iyer@research.iiit.ac.in\\

    \textsuperscript{\rm 2}
    IBM Research, India,
    \{arvagarw,harshitk\}@in.ibm.com

}

\begin{document}
\maketitle
\begin{abstract}
Ontology Alignment is an important research problem that finds application in various fields such as data integration, data transfer, data preparation etc. State-of-the-art (SOTA) architectures in Ontology Alignment typically use naive domain-dependent approaches with handcrafted rules and manually assigned values, making them unscalable and inefficient. 
Deep Learning approaches for ontology alignment use domain-specific architectures that are not only in-extensible to other datasets and domains, but also typically perform worse than rule-based approaches due to various limitations including over-fitting of models, sparsity of datasets etc. In this work, we propose VeeAlign, a Deep Learning based model that uses a dual-attention mechanism to compute the contextualized representation of a concept in order to learn alignments. By doing so, not only does our approach exploit both syntactic and semantic structure of ontologies, it is also, by design, flexible and scalable to different domains with minimal effort. We validate our approach on various datasets from different domains and in multilingual settings, and show its superior performance over SOTA methods.
\end{abstract}

\input{body/intro.tex}

\input{body/method.tex}

\input{body/experiments.tex}

\input{body/relatedwork.tex}
\section{Conclusion}
In this paper we have presented a general purpose ontology alignment method that does not require any external or background knowledge. The method is based on a deep learning architecture where context is modeled explicitly, by first dividing it into different categories based on its relationship with the concept, and then applying a novel dual attention method. The dual attention helps focus on the parts of the context which are most important for the alignment. Our experiments on several datasets from two languages show that the method outperforms the state of the art method by a significant margin. Our ablation study examining the effect of context splitting and dual attention show that these are indeed the factors behind the performance improvement.

\bibliography{refs}

\input{body/appendix}

\end{document}

%% file: body/intro.tex
\section{Introduction}
Ontologies form an integral part of information organization and management. An ontology~\cite{gruber1993translation} is a formal description of knowledge defined using a set of concepts and relations between them. 
Different organizations have different information requirements, and therefore, they follow different nomenclatures (objects and properties) for defining their requirements, resulting in different ontologies for the same underlying data.
As a consequence, to integrate and migrate data among applications, it is crucial to first establish correspondences (or mappings) between the vocabularies of their respective ontologies. Ontology Alignment constitutes the task of establishing correspondences between semantically related elements (i.e. classes and properties) from different ontologies. 

Ontology alignment task has been extensively studied in the last several years, and the solutions have ranged from simple rule based systems~\cite{aml2013,jiang2016ontology} to ones incorporating external knowledge~\cite{hertling2012wikimatch,algergawy2011clustering}, and the most recent ones use sophisticated deep learning based systems~\cite{huang2007ontology, deepalignment2017,wang2018ontology,jimenez2020dividing}. 
Among all the methods, AgreementMarkerLight (AML)~\cite{aml2013} has been a consistent top performer on several ontology alignment tasks organized by OAEI\footnote{\url{http://oaei.ontologymatching.org/}}. 
The Ontology Alignment Evaluation Initiative (OAEI)~\cite{oaei2011} has played a key role in the benchmarking of different ontological alignment systems by facilitating their comparison on the same basis and the reproducibility of the results. 
While AML is one of the best performing systems across different tracks in OAEI, it uses handcrafted rules with manually assigned weights and string similarity algorithms along with domain specific knowledge to discover concept alignments. This kind of approach, while useful, has some obvious limitations. 
Firstly, using string similarity algorithms with minimal focus on context does not address semantic relatedness. 
Secondly, for every pair of ontologies, a new set of rules and weights may need to be defined, which is often a laborious and time consuming process, thus adversely affecting scalability. Deep Learning(DL) based systems have also been used for the ontology alignment task~\cite{deepalignment2017,wang2018ontology}. However, not only do these approaches typically perform worse than rule-based systems, they are also very domain-dependent, with extensive dependability on background knowledge, which, in turn, affects scalability. One of the primary reasons DL architectures use external background knowledge is because of there being a lack of usable training data when it comes to the ontology alignment task. Classification datasets for ontology alignment typically suffer from severe class imbalance and data sparsity, since the number of ground truth alignments is usually several orders smaller than the number of non-alignments. For` example, the conference dataset \cite{conferenceOAEI2014} used for experimentation in this paper has $305$ similar concept pairs and $122588$ dissimilar concept pairs. This data sparsity issue also leads to most standard DL architectures over-fitting and performing poorly. The challenge, therefore, is two fold: a) to use a generic, domain-independent approach to build a training dataset, based solely on the intrinsic semantic and structural information encoded in ontologies with no requirement of external knowledge, and b) to train a model on this dataset that strikes the right balance between model expressivity (which is minimal in rigid rule-based systems) and model complexity (which leads to overfitting).

\begin{figure}[tbh!]
\centering
  \includegraphics[scale=0.40]{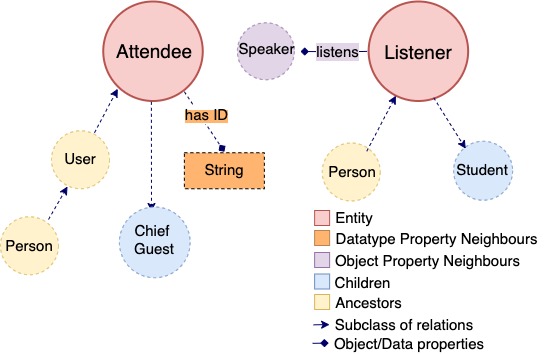} 
\caption{An Example illustrating concept alignment and dependency on the surrounding context}
\label{fig:arch}
\end{figure}

Despite the significant research, ontology alignment still remains a challenging task. Figure~\ref{fig:arch} provides an illustration highlighting this challenge.
The task is to determine the alignment between the concept \textit{Attendee} in Ontology-1 and the concept \textit{Listener} in Ontology-2. Current approaches which work on finding the concept similarity or the neighborhood similarity will fail to capture the alignment between these two concepts since neither of them have high similarity. While there is a common term  (i.e. \textit{Person}) between the contexts of the two concepts, there are several terms which are not similar. This example shows that not only is it important to consider the context, it is also important to model the context in such a way such that the relevant portions of the context have higher weights than the irrelevant portions. In this particular example, ancestor nodes should be given higher weights than children and neighbours connected by object and datatype properties.
Modelling a concept's context in a principled manner is one of the limitations of the existing methods that we address in this work. Note that, a concept in an ontology comes along with rich and diverse context (parent, children, properties); and, it is important that an alignment method is able to make use of it in an effective manner. Among existing methods, AML uses hand crafted structural similarity to include contextual significance, while both the deep learning systems,  DeepAlign~\cite{deepalignment2017} and OntoEmma \cite{wang2018ontology}, do not use ontological context at all.

In this paper, we propose an ontology alignment method, referred to as VeeAlign, that computes mapping between two concepts driven by its context. Our method includes a novel way of modelling context, where the context is split into multiple facets based on the type of neighborhood. More specifically, we divide the context based on its relationship with the central concept such as ancestors, children, object properties and data properties. Such a multi-faceted context, however, poses a new challenge i.e., some of these facets context shall include paths while others will have only neighbouring nodes. In order to deal with this challenge, we propose a dual attention mechanism that comprises of path level attention and node level attention. The path level attention helps find the most important path among all the available paths, whereas the node level attention finds nodes in the path that have the most influence on the central concept alignment. 
The main contributions of this paper are as follows:
\begin{itemize}
\item We model the task of ontology alignment to determine the similarity between two concepts driven by their context. We introduce the notion of multi-faceted context, and model it using a novel attention mechanism, i.e. dual attention.
\item We show through an ablation study the effect of dual attention over single attention and no attention, and the effect of different facet types on model performance
\item We evaluate the proposed model on four datasets, Conference, Lebensmittel, Freizeit, and Web Directory, and show that using the context improves the performance of ontology alignment task, in particular recall of positive alignments.
We choose these datasets to demonstrate the applicability of our approach to diverse data sources in terms of language, domain, and numbers of concepts and concept alignments.
\end{itemize}

%% file: body/method.tex
\section{Approach}
This section presents details of VeeAlign, a Deep Learning based ontology alignment system, that computes contextualized representation of concepts as a function of not just its label, but the multi-faceted neighbours that surround it. In other words, the context is divided into multiple facets based on the relationship between the concept and its neighbours, and then a contextual vector is computed using a dual attention mechanism. This helps the model compute a contextualised representation of a concept, which is later used to compute alignments.

\subsection{Preliminaries}
VeeAlign is a Deep Learning based supervised model trained on positive and negative alignments. Let $O^s$ and $O^t$ be the respective source and target ontologies with $\{e^s_1,e^s_2,\ldots e^s_m\}$ and $\{e^t_1,e^t_2,\ldots e^t_n\}$ be the corresponding elements in the respective ontologies. An ontology consists of different kind of elements such as classes, subclasses, datatype properties, and object properties. In our problem formulation, we consider all these elements i.e., we discover correspondences between elements from one ontology to the respective elements in another ontology. Ontology alignment in its most general form involves finding different kinds of relationships between elements, including complex relationships such as transformation~\cite{thieblin2019survey} or inference~\cite{zhou2018journey}. The focus of this work is to discover the equivalence relationship between elements, primarily because equivalence relations is of the most interest to the community. A formal definition of the ontology alignment task is as following:
\theoremstyle{defn}
\begin{defn}
 Given a source ontology $O^s$ and a target ontology $O^t$, each consisting of a set of elements, the goal of the ontology alignment task is to find all semantically equivalent pairs of elements, i.e. $\{(e^s,e^t) \in O^s \times O^t : e^s \equiv e^t \}$, where $\equiv$ indicates semantic equivalence.
\end{defn}
We now describe the methodology for finding semantically equivalent concepts i.e. classes and subclasses from the given source and target ontologies.

\subsection{Concept Representation}
\begin{figure}[!htb]
	\begin{center}
      \includegraphics[width=0.5\textwidth, angle =0]{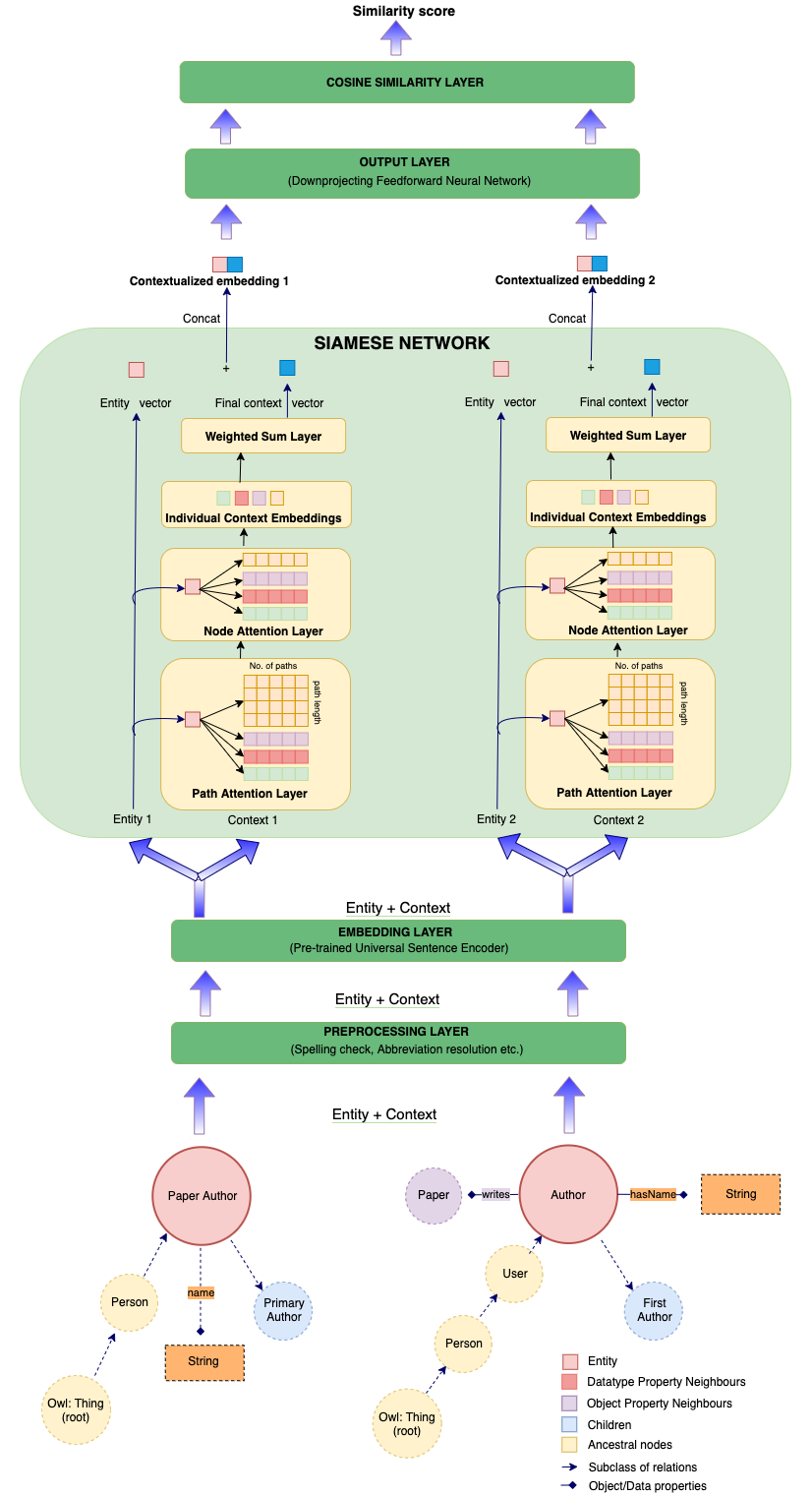}
      \caption{\small VeeAlign Architecture}
    \label{fig:model}
    \end{center}
\end{figure}

The deep learning architecture for VeeAlign is described in Figure~\ref{fig:model}. Since VeeAlign is a supervised model, it requires training data in the form of positive and negative alignment pairs. 
In other words, given a source and target ontology pair as input, we are given ground truth alignment pairs with their labels, i.e. for each $(e^s,e^t) \in O^s\times O^t$, we have $L(e^s,e^t) = 1 $ when $e^s \equiv e^t$, and 0 otherwise. 
For concept alignment, the input to the VeeAlign model are candidate concept pairs $(c^s,c^t)$ along with their labels. Given such a pair as input, a naïve approach could be to find the distributed representations of the elements in the pair and use them for similarity computation. Furthermore, one can use the additional information associated with the concept such as synonym information, description etc for similarity computation. VeeAlign does use label embeddings, but the key difference lies in its method of capturing the context and computing contextualized concept representation, which not only exploits the semantic but also the syntactic and structural nature of ontologies.

\subsection{Context Representation}
We believe that context plays a critical role in alignment, therefore, it is important to model the context in a principled manner. Note that, an ontology consists of concepts along with the relationships among concepts, such as "parent-child" subclass-of relationships, datatype properties, object properties etc. VeeAlign is based on computing the distributed representations of both the concept and its context, which are then concatenated and used to compute the probability score of alignment. 
For a concept $c_i$, let $u_i$ be the distributed representation obtained using Universal Sentence Encoder \cite{cer2018universal}.
%
In VeeAlign, the neighboring concepts connected to a concept forms its context. 
Each neighboring concept has a role and has its own influence on the concept alignment, therefore we categorize neighboring concepts into four categories: ancestor nodes, child nodes, nodes connected through a datatype property and nodes connected through an object property. 


 Sifting through several ontologies and their reference alignments, we observed that two concepts align not just based on their one-hop neighbours, but also on the basis of similarity of "ancestral nodes". In other words, while comparing two concepts, we consider not just their immediate parents, but also the ancestral nodes that lie all the way from the current node to the root node, i.e in the "lineage paths". We thus enumerate all the lineage paths from the current concept to the root and use them for alignments. Let $a_1, a_2,\ldots$ be all the lineage paths for a given concept.  Each lineage path consists of multiple nodes arranged in a sequence, i.e.  $a_j = [c_{a_j1}, c_{a_j2}, \ldots, c_{a_jt} ]$. 

The child nodes consist of concepts which are connected to the current concept using a direct \textsc{Subclass-of} relationship. In order to follow a consistent terminology, we also represent them as a path, however all the paths will have only one node. The contextual concepts linked to the current node through datatype and object properties are represented in the same manner as child concepts, i.e. we only consider immediate one hop concepts linked to the current concept through either of the properties.
\subsection{Dual Attention}
Attention~\cite{Bahdanau2015, paulus2017deep} in deep learning can be broadly interpreted as a vector of weights denoting relative importance. For this task, attention computes weight for neighboring concepts that influence the central concept's alignment. The higher/lower the weight of a concept in a weight vector, the higher/lower is its influence in the central concept alignment computation. 
The dual attention consists of two attentions, one is at the path level, referred to as Path-level attention, and the other is at the node level, referred to as Node-level attention. The goal is to assign a higher weight to the most influential path using Path-level attention. And, within the most influential path assign higher weights to those nodes that are the most influential. The influence of nodes in a path is represented as a weight vector, representing its contribution on the central concept's alignment.
\paragraph{Path-level Attention}
The path level attention aims to find the most important paths for each category. This involves computing the attention weight of each node in each path with respect to the main concept.
Among the 4 different path types, let us first consider lineage paths, i.e. paths containing ancestral nodes. For the concept $c_i$, let $a_j = [c_{a_j1}, c_{a_j2}, \ldots, c_{a_jt} ]$ be one of its lineage paths. The attention weights for each node along the path are calculated as follows:
\begin{equation}
\small
  w_{ia_jk} = u_i^T u_{a_jk}
  \label{eq:path-aw}
\end{equation}
After computing the weights for each node in each lineage path, they are added and normalized to get the relative importance of a path as follows:

\begin{align}
\small
    \begin{split}
        w_{ia_j} &= \sum_{k} w_{a_jk} \\
        w_{ia_j} &= \frac{e^{w_{ia_j}}}{\sum_k e^{w_{ia_j}}} 
    \end{split}
\label{eq:path-att}
\end{align}
Once the relative importance of each path is computed, the next step involves obtaining a unified path representation as a weighted average of all the paths, which is computed by taking the linear combination of all the paths.
Let $w_{ia_1}, w_{ia_2}, w_{ia_t} \ldots$ be the relative importance of all the lineage paths, then
\begin{equation}
\small
   R_{iak} = \sum_j w_{ia_j} u_{ia_jk} 
\end{equation}
where $R_{iak}$ is the representation of a node after combining the representations of the parent nodes from different ancestor paths. The final path is a sequence of node representations i.e. $R_{ia} = [R_{ia1}, R_{ia2}, \ldots R_{iat}]$,
where $t$ is the maximum length of path computed over all ancestor paths. 
\paragraph{Node-level Attention}
As explained earlier, each node in the path contributes towards the central concept's alignment in a proportion relative to its importance, which is determined by the node-level attention. Thus, we apply the second level of attention which ensures that all nodes in the path are combined according to their importance to the central concept. We first compute the attention weights as follows:
\begin{equation}
\small
   \begin{split}
        w_{iak} &= u_i^T R_{iak} \\
        w_{iak} &= \frac{e^{w_{iak}}}{\sum_k e^{w_{iak}}}
    \end{split}
\end{equation}
These attention weights are used to take a weighted linear combination of the node embeddings available in the path embedding $R_{ia}$.
\begin{equation}
    F_{ia} = \sum_k{\theta_{k} \, w_{iak} \, R_{iak}}
\end{equation}
where $F_a$ is the final representation of the ancestors category of the context. $\theta_k$ are trainable parameters introduced to provide importance to each node based on their distance from the central concept. This is driven by the intuition that the immediate ancestors play a more important role in alignment than the distant ones.

\paragraph{Training Layer}
We follow a unified dual attention approach to compute a representation of the context composed of parent nodes, child nodes, nodes connected through datatype properties and object properties. 
The computations for learning representation of parent nodes and other three types of nodes are mostly same, except that, for the other three types there is no notion of path. 
We only have one-hop neighbors so we consider each one-hop neighbour as a path of length one. Next, we apply path-level attention to obtain a unified weighted representation of all the one-hop neighbours. This unified path has only one node, so we skip node-level attention and consider this unified representation as context.


This gives the representations $F_o, F_d, F_c$ corresponding to the nodes connected through the object properties, connected through the datatype properties, and those that are child nodes. We again take a weighted linear combinations of these representations to get the final representation of the context i.e.,


\begin{equation}
    \begin{split}
        v_i &= w_a F_{ia} + w_o F_{io} + w_h F_{ih} + w_d F_{id} \\
            & s.t. \quad w_a + w_o + w_h + w_d = 1
    \end{split}
\end{equation}
This context representation $v_i$ is concatenated with the central concept representation $u_i$, and the combined representation is input to a linear layer for dimensionality reduction in a lower dimension space, as follows:
\begin{equation}
\small
f(c_i) = W * [u_i, v_i]    
\label{eq:finalrep}
\end{equation}
Here, $f(c_i)$ is the final representation of the concept $c_i$. 
For the property alignment, we do not use context and simply take the representation of the names associated with the properties. For a given property $p^s$, we denote by $g(p^s)$  the representation provide by the embedding layer.
Since a candidate alignment pair consists of elements (concepts or properties) from both source and target ontologies, we perform the aforementioned computations for both source and target elements by passing both through a Siamese Network~\cite{bromley1994signature} (which encompasses all the aforementioned attention layers) and then computing the confidence score of the alignment by taking a cosine similarity between the two contextualized representations, i.e. 
\begin{equation}
\small
    \begin{split}
    H(c^s, c^t) &= \cos (f(c^s), f(c^t)) \\
    H(p^s, p^t) &= \cos (g(p^s), g(p^t))
    \end{split}
\label{eq:cos-sim}
\end{equation}
where $(c^s, c^t)$ and $(p^s, p^t)$ are the concept and properties pairs respectively, which we will now denote by elements pair. 
Finally, an element pair $(e^s,e^t)$ is considered a positive alignment when the similarity score is more than a threshold, i.e. $\hat{L}(e^s,e^t) = 1$ when $H(e^s, e^t) > \Theta$ and 0 otherwise.
For the training, we use mean squared error computed as following:
\begin{equation}
\small
    \mathcal{L} = \frac{1}{N}\sum_{(e^s, e^t) \in O^s \times O^t} \Big(H(e^s, e^t) - L(e^s, e^t) \Big)^2
\label{eq:mse}
\end{equation}
where $H(e^s,e^t)$ is obtained using equation\eqref{eq:cos-sim}, and $N$ is total number of training examples. $L(e^s, e^t)$ denotes the ground truth label which is 1 if $e^s \equiv e^t$ and 0 otherwise.

%% file: body/experiments.tex
\section{Experiments}
\label{sec:exp}
In this section, we provide details of the experiments, i.e. the datasets used, baseline models, experimental setup, results and their analysis including ablation study. 

\subsection{Datasets}
We evaluate the performance of our model on four benchmark datasets used in several prior studies for the ontology alignment task (\cite{oaei2011} \cite{peukert2010comparing}). Table \ref{table:concept-distriution} shows the number of concepts in each ontology along with the total number of ground truth positive alignments for entire dataset\footnote{Alignments are meaningful only for an ontology pair, not for a single ontology, therefore we provide the total number of positive alignments available in the entire dataset.}.

\begin{table}[!htb]
\small
\begin{tabular}{|l|l|l|p{2cm}|}
\hline
Dataset & Ontology & \#Concepts & \#Ground Truth Alignments \\ \hline
\multirow{7}{*}{conference} & cmt & 29 & \multirow{7}{*}{305} \\ \cline{2-3}
 & conference & 59 &  \\ \cline{2-3}
 & confOf & 38 &  \\ \cline{2-3}
 & edas & 103 &  \\ \cline{2-3}
 & ekaw & 73 &  \\ \cline{2-3}
 & iasted & 140 &  \\ \cline{2-3}
 & sigkdd & 49 &  \\ \hline
\multirow{2}{*}{Lebensmittel} & Google & 58 & \multirow{2}{*}{32} \\ \cline{2-3}
 & web & 52 &  \\ \hline
\multirow{2}{*}{Freizeit} & dmoz & 70 & \multirow{2}{*}{67} \\ \cline{2-3}
 & Google & 66 &  \\ \hline
\multirow{4}{*}{Web directory} & dmoz & 745 & \multirow{4}{*}{2051} \\ \cline{2-3}
 & Google & 727 &  \\ \cline{2-3}
 & web & 417 &  \\ \cline{2-3}
 & Yahoo & 1132 &  \\ \hline
\end{tabular}
\caption{Concept and alignment distribution for each dataset}
\label{table:concept-distriution}
\end{table}

\begin{table*}[tbh]
\small
\centering
\begin{tabular}{|l|c|c|c||c|c|c||c|c|c||c|c|c|}
\hline
Dataset & \multicolumn{3}{c|}{VeeAlign} & \multicolumn{3}{c|}{AML} & \multicolumn{3}{c|}{LogMap2} & \multicolumn{3}{c|}{DeepAlign} \\ \hline
 & P & R & F & P & R & F & P & R & F & P & R & F \\ \hline
Conference & 0.774 & 0.741 & \textbf{0.748} & 0.802 & 0.651 & 0.7 & 0.821 & 0.654 & 0.701 & 0.631 & 0.586 & 0.567 \\ \hline
Lebensmittel & 0.641 & 0.562 & \textbf{0.596} & 0 & 0 & nan & 1 & 0.3 & 0.437 & - & - & - \\ \hline
Freizeit & 0.788 & 0.985 & \textbf{0.874} & 0 & 0 & nan & 0.925 & 0.747 & 0.821 & - & - & - \\ \hline
Web directory &  0.468	& 0.695	& 0.559  & - & - & - & 0.778 & 0.542 & \textbf{0.638} & - & - & - \\ \hline
\end{tabular}
\caption{Performance comparison of VeeAlign with the baseline methods. ``-" means that the results could not be obtained either due to inapplicability of the method on the dataset or it not being able to finish under a reasonable time limit. $P=0$ and $R=0$ mean that the algorithm did not output any alignments.}
\label{tab:mainresults}
\end{table*}

\begin{table}[tbh]
\small
 \centering
\begin{tabular}{|l|c|c|c|c|}
\hline
 & VeeAlign & AML & LogMap2 & DeepAlign \\ \hline
cmt-ekaw & 0.545 & 0.6 & 0.632 & \textbf{0.666} \\ \hline
cmt-sigkdd & 0.8 & 0.818 & \textbf{0.957} & 0.78 \\ \hline
conf-sigkdd & 0.69 & 0.692 & \textbf{0.759} & 0.615 \\ \hline
edas-iasted & \textbf{0.667} & \textbf{0.667} & 0.519 & 0.47 \\ \hline
ekaw-iasted & \textbf{1} & 0.737 & 0.7 & 0.46 \\ \hline
cmt-confOf & \textbf{0.643} & 0.583 & 0.455 & 0.364 \\ \hline
cmt-iasted & \textbf{0.889} & 0.8 & \textbf{0.889} & 0.615 \\ \hline
\end{tabular}
\caption{F-score on individual ontology pairs from Conference dataset}
\label{tab:conf_detail}
\end{table}


\begin{itemize}
  \item Conference~\cite{conferenceOAEI2014}: The OAEI Conference dataset consists of 16 ontologies from the conference organization domain with ground truth alignments provided for 7 of them, resulting in 21 ontology pairs. 
  
  \item Lebensmittel~\cite{peukert2010comparing}: This dataset consists of ontologies from Food domain. The  concepts and ground truth alignments are extracted from web directories of Google and the web. Both ontologies are in the German language.
  
  
  \item {Freizeit}~\cite{peukert2010comparing}: Similar to Lebensmittel, this dataset is also in German, and consists of concepts and ground truth alignments pairs extracted from web directories of dmoz and Google, related to online shopping in the Leisure domain.
  
  \item {Web directory}~\cite{massmann2008evaluating} :  This dataset, in German, contains relatively larger ontologies. 
  The four ontologies in this dataset consist of concepts related to online shopping websites that are extracted from dmoz, Google, web and Yahoo web directories.

\end{itemize} 

The datasets were selected in order to be able to fairly evaluate a general purpose domain-independent ontology alignment system that does not use any background knowledge and is also suitable for different languages. Language is an important consideration in ontology alignment, as several applications of the ontology alignment problem such as data integration, data transfer require system to be operable for multiple languages.


\subsection{Hyperparameters}

In our implementation, we use the following hyperparameters optimized through grid-search. The word vectors for each concept were initialized with 512-dimension Universal Sentence Encoder (USE)~\cite{cer2018universal} for the conference dataset and its multilingual variant \cite{yang2019multilingual} for the 3 German-language datasets respectively. The model was converged using MSE loss and Adam optimizer with a learning rate of 0.001, after training for 50 epochs with a batch size of 32. We experimented with another variation of the model where one-hop properties and children neighbors are represented as one path of length $L$ created with random sequence, as opposed to $L$ paths of length 1. For obtaining a unified path representation, we experiment with weighted sum and max pooling, and report the best results. Finally, the dimension of the down-projecting output layer was set to 300. All randomizations including PyTorch, Numpy is done using 0 as the seed. More details on the experimental setup including computing infrastructure is provided in Appendix A.

\subsection{Data Preprocessing and Evaluation Methodology} 
Our model takes the positive and negative alignment pairs as input, and since we are only given positive alignment pairs, we construct negative alignment pairs by first creating all the possible pairs from input ontologies and selecting the ones which are not part of the ground truth alignments. The data consisting of the positive and negative pairs is split
into training, validation and test sets using the K-fold "sliding window" method of evaluation. For the conference dataset, we do 7-fold cross validation, so out of 21 ontology pairs, 6 folds (18 pairs) are used for training, 1 fold for training and validation (2 pairs from 1 fold for validation and 1 pair for testing). While evaluating on Lebensmittel, Freizeit and Web directory datasets we perform 5-fold cross validation, in which 70\% of the concept-level alignments are used for training, 10\% for validation and 20\% for testing. Since the conference dataset consists of 21 pairs of small ontologies, we split them at the ontology-pair level. Whereas, Lebensmittel, Freizeit and Web Directory datasets consist of 1, 1 and 6 pairs of ontology alignments respectively, we split at the concept-pair level in order to obtain reasonable amounts of training data for facilitating the training process. In each fold, during training, we over-sample the positive alignments in order to maintain a 1:1 ratio between positive and negative alignments. By doing so, we address the common problem of class imbalance in ontology alignment problems, since the total number of possible pairs is typically several magnitudes larger than the number of ground truth alignments. We use the validation set for various hyper-parameter optimization including finding the optimal threshold used during testing. We use precision, recall and F1-score of the positive class as our evaluation metric.

\subsection{Results and Discussion}
We present the results of our experiments in Tables~\ref{tab:mainresults} and \ref{tab:conf_detail}. Table~\ref{tab:mainresults} shows the precision(P), recall(R) and F1-score(F) for our algorithm in comparison to the baseline algorithms, AML\cite{aml2013}, LogMap2 \cite{jimenez2012large} and DeepAlign \cite{deepalignment2017}. Note that, we were not able to run all baselines for all datasets. In particular, AML timed-out on the Web directory dataset because since it contains rules that have only been adapted for the OAEI tracks, it defaults to a variety of string similarity matches which take considerable time to compute due to combinatorial nature of unoptimized string computations. 
Also, DeepAlign could not be run on German datasets because of the unavailability of the synonyms and antonyms in German language. We were able to run all baselines on the Conference dataset. From the results in Table 1, we observe that VeeAlign significantly outperforms on 4 out of 3 datasets on the F1-score metric. In Conference dataset, VeeAlign has 4.7\% point improvement in comparison to the AML (the second best performing model).  Whereas, on  Lebensmittel and Freizeit, VeeAlign achieves 15.9\% and 5.3\% points improvement respectively in comparison to the LogMap2 method. An important observation from these results is that both AML and LogMap2 have very high precision on all datasets which is justifiable given that these are manually drafted rule based systems. For any rule-based system, high precision output is expected since it is very easy to draft rules for certain cases and get them right, however getting high recall is challenging. In contrast, the statistical methods (VeeAlign and DeepAlign) provide a balance between recall and precision. When compared to DeepAlign, VeeAlign significantly outperforms it on both recall and precision. Table~\ref{tab:conf_detail} shows detailed comparison of the baseline methods with VeeAlign on seven different ontological pairs from the conference dataset. 
From these results, we observe that VeeAlign preforms better on 4 out of 7 pairs, whereas LogMap2 and DeepAlign performs better on 2 and 1 pairs, respectively. 
These results again shows the limitations of the AML and LogMap2 methods which have been the best performer on benchmarks datasets, however, when compared to statistical methods, they have rather inferior performance. 
Another observation from these results is the performance consistency of DeepAlign, especially when compared to other statistical method, i.e. VeeAlign. Note that, the DeepAlign has rather inconsistent performance, i.e. on some ontology pairs, it performs exceptionally well while on others not so good. For DeepAlign, F2-score ranges from 28\% to 76\% whereas for VeeAlign, it is between 59\% and 100\%. 




\begin{table}[tbh]
\small
\centering
\begin{tabular}{|l|c|c|c|}
\hline
 & P & R & F \\ \hline
No Context & 0.775 & 0.608 & 0.670 \\ \hline
Context   + single attention &  0.678	& 0.728	&0.697 \\ \hline
Context   + dual attention &  0.774	& 0.741	& 0.748 \\ \hline
\end{tabular}
\caption{Effect of incorporating context using single and dual attention}
\label{tab:ablation_attn}
\end{table}

\begin{table}[tbh]
\small
\centering
\begin{tabular}{|l|c|c|c|}
\hline
Context & P & R & F \\ \hline
Parents &  0.747	& 0.707	& 0.719  \\ \hline
Children &  0.634	&. 0.75	& 0.678  \\\hline
Object   Properties &  0.647	& 0.74	& 0.681  \\\hline
Data   prroperties &  0.64	& 0.75	& 0.681  \\\hline
All combined &  0.774	& 0.741	& 0.748\\\hline 
\end{tabular}
\caption{Effect of different types of context}
\label{tab:ablation_context}
\end{table}

\paragraph{Effect of Context and Dual Self Attention}
We perform an ablation study and analyze the effect of different layers on model performance, in particular when there is no context, using context with single attention (not using path level information) and and using context with dual attention i.e., using both path and node level information. The results are shown in Table~\ref{tab:ablation_attn}. These results indicate that the dual attention is an improved model over single attention which is an improved model over when there is no context, demonstrating the efficacy of context and of modeling it using dual attention. It is worth noting that modeling the context using dual attention gives a significant improvement in recall and F-score, primarily because the model considers a richer set of information for alignment, but at the same time does not reduce precision because information is used based on its importance for the alignment task.

\paragraph{Effect of Context Type}
Another dimension to analyze the model performance is along the lines of using different types of context information i.e., parent, children, data properties and object properties. The results from this analysis are show in Table~\ref{tab:ablation_context}. These results indicate that parents are the most useful type of context information whereas the children are of the least. However the best alignment results are obtained when we combined all four types of context, this combination gives us a performance improvement in both recall and precision.

%% file: body/relatedwork.tex
\section{Related Work}
There has been a large body of work on the ontology alignment problem~\cite{euzenat2007ontology, otero2015ontology, niepert2010probabilistic, schumann2015minimizing}, primarily driven by the OAEI (Ontology Alignment Evaluation Initiative). OAEI has been conducting ontology alignment challenges since 2004 where multiple datasets belonging to different domains are released along with a public evaluation platform to evaluate different systems. Among all the systems submitted to the challenge, two systems have consistently outperformed the others. 

The first is AgreementMakerLight (AML)~\cite{aml2013} which uses a combination of various matching algorithms called matchers such as lexical matcher reflecting the lexical similarities between the entities, structural matcher which compares ontology concepts or their instances based on their relationships with other concepts or instances. The recent AML system~\cite{santos2015ontology} also include a repair algorithm  that minimizes the incoherence of the resulting alignment and the number of matches removed from the input alignment. The second best performing system is LogMap2~\cite{logmap2020} that is specifically designed to align large scale ontologies. The system works in an iterative manner, starting from the initial anchors, alternates mapping repair and mapping discovery steps.  Both of these systems have been heavily engineered over the years to give the best performance on the datasets and domains available in OAEI. However when applied on the datasets other than OAEI, their performance is rather less impressive. These observations have been supported by the experimental study as well. Furthermore these systems are highly domain specific requiring specialized knowledge in terms of lexicons.



The ontology alignment community has only recently started to look into statistical methods, in particular, Deep Learning based methods where models are trained on the given input data. However, these systems are typically even more domain-specific, and require background knowledge in order to train. For instance, a recent work by \cite{wang2018ontology} presents a neural network based ontology alignment system for the Biomedical domain. The idea is to enrich entities in an ontology with aliases from the ontology, definitions from Wikipedia and context from background knowledge sources, and use this additional information for ontology alignment. Similarly, DeepAlign\cite{deepalignment2017} too requires synonyms and antonyms extracted from external sources such as WordNet and PPDB in order to refine the word vectors using synonymy and antonymy constraints, which are subsequently used for alignment. Such dependencies can have severe limitations when dealing with, say, multilingual ontologies, or small generic ontologies where no background knowledge is available. In addition, they typically perform worse than rule-based systems, and even more so when tested on other domains.

In contrast to these methods, VeeAlign does not require any external background knowledge. It completely relies on the semantic and structural information encoded in an ontology, in particular the contextual information available with entities to learn better representations.

%% file: body/appendix.tex
\appendix
\label{appendix:A}